\documentclass[sigconf]{aamas} 

\usepackage{forest}

\usepackage{balance} 

\setcopyright{ifaamas}
\acmConference[AAMAS '21]{Proc.\@ of the 20th International Conference on Autonomous Agents and Multiagent Systems (AAMAS 2021)}{May 3--7, 2021}{London, UK}{U.~Endriss, A.~Now\'{e}, F.~Dignum, A.~Lomuscio (eds.)}
\copyrightyear{2021}
\acmYear{2021}
\acmDOI{}
\acmPrice{}
\acmISBN{}

\acmSubmissionID{???}

\title[FOL Neural MCTS]{First-Order Problem Solving through Neural MCTS based Reinforcement Learning}

\author{Ruiyang Xu}
\affiliation{
  \department{Khoury College of Computer Sciences}
  \institution{Northeastern University, Boston}}
\email{ruiyang@ccs.neu.edu}

\author{ Prashank Kadam}
\affiliation{
  \department{Khoury College of Computer Sciences}
  \institution{Northeastern University, Boston}}
\email{kadam.pr@northeastern.edu}

\author{Karl Lieberherr}
\affiliation{
  \department{Khoury College of Computer Sciences}
  \institution{Northeastern University, Boston}}
\email{lieber@ccs.neu.edu}

\DeclareMathOperator*{\argmax}{arg\,max}

\begin{abstract}
The formal semantics of an interpreted first-order logic (FOL) statement can be given in Tarskian Semantics or a basically equivalent Game Semantics. The latter maps the statement and the interpretation into a two-player semantic game. Many combinatorial problems can be described using interpreted FOL statements and can be mapped into a semantic game. Therefore, learning to play a semantic game perfectly leads to the solution of a specific instance of a combinatorial problem. We adapt the AlphaZero algorithm so that it becomes better at learning to play semantic games that have different characteristics than Go and Chess. We propose a general framework, Persephone,  to map the FOL description of a combinatorial problem to a semantic game so that it can be solved through a neural MCTS based reinforcement learning algorithm. Our goal for Persephone is to make it tabula-rasa, mapping a problem stated in interpreted FOL to a solution without human intervention.   
\end{abstract}


\newcommand{\BibTeX}{\rm B\kern-.05em{\sc i\kern-.025em b}\kern-.08em\TeX}

\begin{document}


\pagestyle{fancy}
\fancyhead{}


\maketitle 


\section{Introduction}
Recent success from AlphaZero \cite{alpha0} sheds light on solving combinatorial games through combining deep RL with Monte Carlo Tree Searching (MCTS). It is known that conventional model-free RL performs poorly on those problems, that is because a combinatorial problem usually implies a large state space with sparse rewards, which causes sample efficiency a challenge to those algorithms. Nevertheless, neural MCTS can largely increase the sample efficiency and improve the performance of solving those problems.

Since the first proposal of neural MCTS \cite{alpha_go,exit}, people have seen a remarkable performance of the algorithm on gameplay and tried to improve this algorithm in various ways. For instance, AlphaZero \cite{alpha0} uses self-play to generate learning data, which makes it becomes self-supervised learning. However, the power of AlphaZero is limited to conventional board games. Later, MuZero \cite{muzero} extended this idea to model-based RL, which makes the algorithm able to play Atari games. At the theoretical level, Grill et al. \cite{Grill2020MonteCarloTS} pointed out that MCTS itself is some kind of regularized policy optimization, which perfectly explained the mysterious PUCT heuristic appeared in the AlphaZero paper. Furthermore, based on their theory, they proposed an optimal heuristic and proved that one has a better performance than AlphaZero's. Their research reveals a deep connection between neural MCTS and RL, which has also been noticed by other researchers, like in \cite{Hamrick2020CombiningQA} where the author uses a single Q-value network to replace the value and policy network in AlphaZero. The policy is dynamically computed from the Q-values through a softmax operator. Performance improvement has also been seen in this variant. Likewise, Guo et al. \cite{ijcai2018-523} use a three-headed neural network (i.e., policy, value, and Q-value) to increase the learning efficiency on Hex. Moreover, recently, Goldwaser et al. \cite{Goldwaser2020DeepRL} adapted neural MCTS to general game playing \cite{genesereth2005general}, however, their current framework can only handle turn-based two-player zero-sum symmetric games.

Neural MCTS's ability to handle games with large state space and sparse reward motivates us to extend its application to another domain, solving combinatorial problems. We should mention that applying RL to solving combinatorial problems has been studied for a period of time \cite{laterre2018ranked,bello2016neural,khalil2017learning,Xu2019LearningSA,Mazyavkina2020ReinforcementLF,cappart2020combining}. In this paper, unlike other methods, we propose a framework, Persephone, which solves FOL-expressible combinatorial problems (first-order problems in short) through a logic semantic-based gamification. We also point out a close connection between neural MCTS and RL to use concepts in RL to improve the algorithm. Specifically, our main contributions are 1. we proposed a framework to map first-order problems to multi-agent MDPs so that solutions can be learned through neural MCTS based RL. 2. We define symmetry and asymmetry for extended form games and exploit the asymmetry in our network designs. 3. We propose using warm-start MCTS, different policy learning strategies, and asymmetric neural network structures to increase the performance. 4. We carry out experiments on different designs and verify the best configuration. Our experimental results show that using asymmetric designs helps increase performance on asymmetric semantic games.


\section{Preliminary}
\subsection{Neural MCTS}
MCTS has been applied to solving combinatorial games for a long time \cite{mcts_survey}, while recently, combining deep neural networks with MCTS showed success in improving solver competence in many practical combinatorial games. The concept of neural MCTS was proposed independently in Expert Iteration \cite{exit} and AlphaZero \cite{alpha0}. In a nutshell, neural MCTS uses the neural network as policy and value approximators. During each learning iteration, it carries out multiple rounds of self-plays. Each self-play runs several MCTS simulations to estimate an empirical policy at each state, then sample from that policy, take a move, and continue. After each round of self-play, the game's outcome is backed up to all states in the game trajectory. Those game trajectories generated during self-play are then be stored in a replay buffer, which is used to train the neural network.

In self-play, for a given state, the neural MCTS runs a given number of simulations on a game tree ,rooted at that state, to generate an empirical policy. Each simulation, guided by the policy and value networks, passes through 4 phases:
\begin{enumerate}
\item{SELECT: }
At the beginning of each iteration, the algorithm selects a path from the root (current game state) to a leaf (either a terminal state or an unvisited state) according to an upper confidence boundary (UCB, \cite{puct,ucb1,uct}). Specifically, suppose the root is $s_0$. The UCB determines a serial of states $\{s_0, s_1, ..., s_l\}$ by the following process:
\begin{equation}
    \begin{split}
    a_{i}&=\argmax_a\left[Q(s_{i},a)+c\pi_\theta(s_{i},a)\frac{\sqrt{\sum_{a'} N(s_{i},a')}}{N(s_{i},a)+1}\right]\\
    s_{i+1}&=\text{move}(s_{i},a_{i})
    \end{split}
    \label{eq:mcts-select}
\end{equation}
It has been proved in \cite{Grill2020MonteCarloTS} that selecting simulation actions using Eq.\ref{eq:mcts-select} is equivalent to optimize the empirical policy $$\hat{\pi}(s,a)=\frac{1+N(s,a)}{|A|+\sum_{a'} N(s,a')}$$ 
where $|A|$ is the size of current action space, so that it approximate to the solution of the following regularized policy optimization problem:
\begin{equation}
\begin{split}
    \pi^*&=
    \argmax_\pi\left[Q^T(s,\cdot)\pi(s,\cdot)- \lambda KL[\pi_\theta(s,\cdot),\pi(s,\cdot)]\right]\\
    \lambda &= \frac{\sqrt{\sum_{a'} N(s_{i},a')}}{|A|+\sum_{a'} N(s,a')}
\end{split}
    \label{eq:mcts-opt}
\end{equation}
That also means that MCTS simulation is an regularized policy optimization \cite{Grill2020MonteCarloTS}, and as long as the value network is accurate, the MCTS simulation will optimize the output policy so that it maximize the action value output while minimize the change to the policy network.
\item{EXPAND: }
Once the selected phase ends at an unvisited state $s_l$, the state will be fully expanded and marked as visited. All its child nodes will be considered as leaf nodes during next iteration of selection.
\item{ROLL-OUT: }
The roll-out is carried out for every child of the expanded leaf node $s_l$. Starting from any child of $s_l$, the algorithm will use the value network to estimate the result of the game, the value is then backed up to each node in the next phase.
\item{BACKUP: }
This is the last phase of an iteration in which the algorithm updates the statistics for each node in the selected states $\{s_0, s_1, ..., s_l\}$ from the first phase. To illustrate this process, suppose the selected states and corresponding actions are
$$\{(s_0,a_0),(s_1,a_1),...(s_{l-1},a_{l-1}),(s_l,\_)\}$$ 
Let $V_\theta(s_i)$ be the estimated value for child $s_i$. We want to update the Q-value so that it equals to the averaged cumulative reward over each accessing of the underlying state, i.e., $Q(s,a)=\frac{\sum_{i=1}^{N(s,a)}\sum_tr_t^i}{N(s,a)}$. To rewrite this updating rule in an iterative form, for each $(s_t,a_t)$ pair, we have:
\begin{equation}
\begin{split}
    N(s_t,a_t)&\leftarrow N(s_t,a_t)+1\\
    Q(s_t,a_t)&\leftarrow Q(s_t,a_t)+\frac{V_\theta(s_r)-Q(s_t,a_t)}{N(s_t,a_t)}
\end{split}
    \label{eq:mcts-backup}
\end{equation}

Such a process will be carried out for all of the roll-out outcomes from the last phase.
\end{enumerate}
Once the given number of iterations has been reached, the algorithm returns the empirical policy $\hat{\pi}(s)$ for the current state $s$. After the MCTS simulation, the action is then sampled from the $\hat{\pi}(s)$, and the game moves to the next state. In this way, for each self-play iteration, MCTS samples each player's states and actions alternately until the game ends, which generates a trajectory for the current self-play. After a given number of self-plays, all trajectories will be stored into a replay buffer so that it can be used to train and update the neural networks.

\subsection{Game-theoretic Semantics}
Game semantics is an approach that rebuilds the logical concepts of game-theoretic concepts. For instance, in the propositional logic, each formula is interpreted as a game between two players. The Proponent is in charge of all the "OR" operators, while the Opponent takes over all the "AND" operators. The game runs recursively on the computational order of the operators. During each move of the game, the current operator owner will choose one of its sides as a subformula, and the game will then continue in that subformula. The game will end when a primitive proposition is achieved, and the Proponent wins the game if the formula evaluates to true; otherwise, the Opponent wins.

Hintikka refined the game semantics and extended it to the model-based first-order logic. To be specific, using Hintikka's game-theoretic semantics approach \cite{semgame}, one can map any first-order logic formula (with models) into object picking and fact testing. A winning strategy for the Proponent/Opponent exists if the underlying logic formula is true/false. Under the game-theoretic semantics, a semantic game is represented as a tuple $(\langle \Psi,\ M \rangle,\ \text{P},\ \text{OP})$, where the underlying formula is $\Psi$ interpreted by the model $M$. $ P $ and $ OP $ denote the game role, namely, who is playing as the Proponent/Opponent. There are 6 types for any first-order logic formula (Tbl.\ref{table:SGRules}):
\begin{enumerate}
    \item For universally quantified formulas, the Opponent takes a move by providing a legal value $x_0$ for the quantified variable $x$. The game continues with a subformula where all the appearances of $x$ are replaced with the particular assignment $x_0$.
    \item For existentially quantified formulas, the Proponent takes a move by providing a legal value $x_0$ for the quantified variable $x$. The game continues with a subformula where all the appearances of $x$ are replaced with the particular assignment $x_0$.
    \item For conjunctive formulas, the Opponent takes a move by picking either side of the subformula from the original formula.
    \item For disjunctive formulas, the Proponent takes a move by picking either side of the subformula from the original formula.
    \item For negated formulas, no moves will happen, while the two players will switch roles. 
    \item For primitive propositions, the true value will be evaluated directly within the given model $M$. This situation is also an indication of the end of the game, where the player who currently plays Proponent wins if the formula evaluates to true; otherwise, the current Opponent player wins.
\end{enumerate}
\newcommand{\raisedChi}{\raisebox{2pt}{$\chi$}}
\begin{table}[ht]
\centering
    \begin{tabular}{|c|c|c|}
        \hline
        
        
        Proposition $\varphi$ & Operation & Subgame \\ \hline
        $\forall x: \Psi(x)$ & OP picks $x_0$ & $(\langle \Psi[x/x_0],\ M \rangle,\ \text{P},\ \text{OP})$ \\
        
        $\Psi \wedge \raisedChi$ &  OP picks $\theta \in \{\Psi, \raisedChi \}$ & $(\langle \theta,\ M \rangle,\ \text{P},\ \text{OP})$\\

        $\exists x: \Psi(x)$ & P picks $x_0$ & $(\langle \Psi[x/x_0],\ M \rangle,\ \text{P},\ \text{OP})$ \\
        
        $\Psi \vee$ $\raisedChi$ &  P picks $\theta \in \{\Psi, $ \raisedChi$ \}$ & $(\langle \theta,\ M \rangle,\ \text{P},\ \text{OP})$\\

        $\neg \Psi$ & N/A & $(\langle \Psi,\ M \rangle,\ \text{OP},\ \text{P})$ \\
        
        $\varphi$ & N/A & N/A\\
        \hline
    \end{tabular}
    \caption{The mapping between a formula $\langle \varphi,\ M\rangle$ and the corresponding semantic game $(\langle \varphi,\ M\rangle,\ \text{P},\ \text{OP})$. In this table, OP stands for Opponent, and P stands for Proponent. M is the model of the formula, which defines all non-logical symbols in the formula. The game ends at an atomic proposition $\varphi$. It is to be noted that the negation switches the role of the two players; namely, strategies for P in a game for $\neg \Psi$ are strategies for OP in the game for $\Psi$.}
    \label{table:SGRules}
\end{table}
\subsection{Asymmetric Extensive Form Games}
In this section, following the definition given by \cite{Selten1983EvolutionarySI, Heller2014StabilityAT}, we extend the concept of the asymmetric game from normal form game to the two-player extensive form game using first-order logic. The asymmetry of an asymmetric game is mainly reflected in two aspects: 1. The players' actions space is asymmetric. Namely, players with different game roles have different action spaces. 2. the final objective of the game roles are asymmetric. Generally speaking, any two-player extensive form game can be abstractly described as the following:
$$G(S,P)=\begin{cases} \mbox{True},\ \mbox{if P wins} \\ \mbox{False},\ \mbox{if P loses}
\\\exists m\in \mathcal{A}_P^S:\ \neg G(\mathcal{T}(S,m),\neg P)
\end{cases}$$
where $G(S,P)$ is a predicate that claims that, given a game state parameter $S$, the current player $P$ will win the game. In a nontrivial case, the current player will pick up the action from the action space $\mathcal{A}_P^S$ and claim that after taking this action through the transition operator $\mathcal{T}(S,m)$, the other player cannot win the game. It is to be noted that the action space is parameterized on the current state and player, which is consistent with any game definition.
\begin{definition}
Given the space of all possible legal game states $\mathfrak{S}$ the space of all possible legal actions $\mathfrak{A}$, an extensive form game is symmetric if there exist two involutions $\varphi:\mathfrak{A}\rightarrow \mathfrak{A}$ and $\psi:\mathfrak{S}\rightarrow \mathfrak{S}$ such that we have the following holds:
\begin{equation}
\begin{split}
    &\forall m\in \mathfrak{A}: \varphi(m)\in \mathfrak{A}\wedge \varphi(\varphi(m))=m\\
    &\forall S\in \mathfrak{S}: \psi(S)\in \mathfrak{S}\wedge \psi(\psi(S))=S\\
    &\forall m\in \mathcal{A}_P^S: \varphi(m)\in\mathcal{A}_{\neg P}^{\psi(S)}\\
    &\forall S\in \mathfrak{S}: G(\psi(S),\neg P)=G(S,P)
\end{split}
    \label{eq:asym}
\end{equation}
\end{definition}
Using the definition above, one can easily judge whether a given extensive form game is symmetric or asymmetric. For instance, Chess is considered symmetric because one can find $\psi$ and $\varphi$ both to be a 'flipping' operator, which makes the current board upside down. That means the White player can move by pretending to be a Black player by rotating the board 180 degrees and flipping the color. On the other hand,  Fox and Geese is asymmetric, because one cannot find any involution $\varphi$ between Fox player's action space and Geese player's action space such that $\psi(S)$ is a legal state in $\mathfrak{S}$. It can also be inferred that all impartial games (hence all Nim games) are symmetric because, by definition, all players share the same action space, so the mappings here can just be identical. Nevertheless, partisan games can either be symmetric or asymmetric based on the game definition.     

The semantic games we are dealing with are considered asymmetric in general because the two roles' action space would be different due to different domains for quantified variables. Such an asymmetry introduces an intrinsic imbalance once the two players play against each other, which causes the game to be easier for one player but harder for the other one. The asymmetric games become a challenge for the original AlphaZero algorithm designed for symmetric board games, like Chess and Go. As a result, AlphaZero can learn on a consistent action space, objective, and player role using the color-flipping trick. The symmetric games make it possible only to learn a single policy that applies to both players. However, separate policies have to be learned for an asymmetric game, adding another layer of complexity to the learning algorithm.


\section{Methodology}
\subsection{Overview of Persephone}
\begin{figure}[h]
  \centering
  \includegraphics[width=\linewidth]{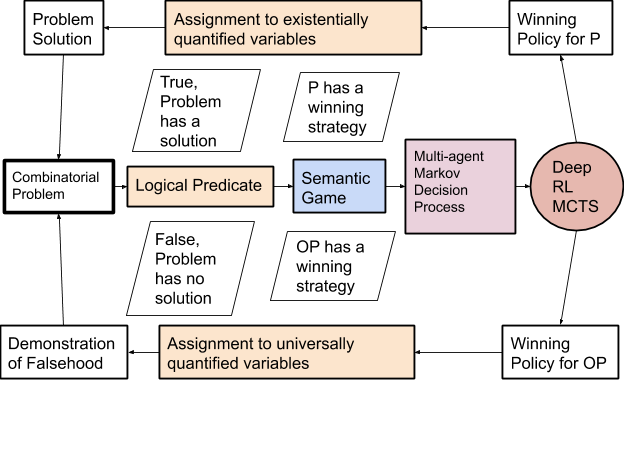}
  \caption{The Persephone Framework.}
  \label{fig:persephone}
  \Description{The Persephone Framework.}
\end{figure}

In this section, we propose a general framework, Persephone, to solve FOL-expressible combinatorial problems. Through game-theoretic semantics, the framework transforms the FOL description of the target problem into a two-player semantic game. The transformed two-player semantic game can then be modeled with Two-player MDPs. After that, a neural MCTS algorithm will be applied to play and learning the game. The algorithm will finally converge to an optimal strategy for the Proponent player if the original problem has an optimal solution. Otherwise, the Opponent player will learn the counter strategy to demonstrate the falsehood of the original problem (Fig. \ref{fig:persephone}). 

Two-player MDPs can be viewed as extensions of MDPs \cite{markovgame}, which, in case of semantic games, can be represented as a tuple $\langle \mathcal{S}_P, \mathcal{A}_P, \mathcal{S}_{OP}, \mathcal{A}_{OP}, \mathcal{R}, \mathcal{T}, \gamma\rangle$ where: $\mathcal{S}_1$ and $\mathcal{S}_2$ are state spaces for each players, and  $\mathcal{A}_1$ and $\mathcal{A}_2$ are action spaces. 
\begin{itemize}
    \item $\mathcal{S}_P$ and $\mathcal{S}_{OP}$ are state spaces for each players, which contains all possible states in a decision problem. In terms of the semantic game, they contain all possible legal game states for each players respectively. 
    \item $\mathcal{A}_P$ and $\mathcal{A}_{OP}$ are action space for each player, which contains all possible actions in a decision problem. In terms of the semantic game, it contains all possible legal moves for each players respectively. 
    \item transition function $\mathcal{T}$ defines the dynamic from one state to another. In a semantic game, we have a transition function $\mathcal{T}: \mathcal{S}_{P/OP} \times \mathcal{A}_{P/OP} \rightarrow \mathcal{S}_{P}\cup S_{OP}$. It is to be noted that a feature of the semantic game is that, depending on the step-wise evaluation of the FOL formula and game-theoretical semantics, the next state can either belong to the same player or change to another player.
    \item Rewards $\mathcal{R}(s,a,s')$, which defines the reward after taking action $a$ in state $s$ and moving to state $s'$. In a semantic game, since the outcome of the game is unknown until the game ends, the rewards are sparse. In other words, we have the following reward function (suppose $s\in \mathcal{S}_{P/OP}$)
    $$\mathcal{R}(s,a,s')=\begin{cases} eval(s'),\ \mbox{if } s' \mbox{is terminal and } s'\in\mathcal{S}_{P/OP} \\ -eval(s'),\ \mbox{if } s' \mbox{is terminal and } s'\in\mathcal{S}_{OP/P} 
    \\0,\ \mbox{otherwise} 
    \end{cases}$$
    $$eval(s)=\begin{cases} 1,\ \mbox{if } s \mbox{ evaluates to True}\\ -1,\ \mbox{if } s \mbox{ evaluates to False} 
    \end{cases}$$
    \item Reward discount factor $\gamma\in (0,1]$, which weighs the importance of future rewards. Typically, the farther the distance of a reward from the current state, the less effective the reward brings to the current decision. In our semantic games, since the reward is sparse, $\gamma$ is set to 1.
\end{itemize}
Since there are two players, solving these MDPs means finding two policies $\pi_P$ and $\pi_{OP}$ such that a Nash equilibrium can be established. As a result, once the learning converged, one of the player can always win the game while the other one is forced to lose the game.
\subsection{Implementation Details}
\subsubsection{State Representation}
The entrance of a semantic game is always a predicate, which can be viewed as a tree. Once evaluating this predicate step-wisely, each node is either a logic operator or a predicate. A predicate indicates a leaf node for the current tree, but also an entrance to another tree. Persephone uses a preorder traversal to identify each node, and hence to vectorize the tree structure uniquely. For instance,
$$HSR(k,q,n):=\exists m\in[1,n):HSR(k-1,q-1,m)\wedge HSR(k,q-1,n-m)$$
can be represented as:
\begin{center}
    \begin{forest} 
[Exists
    [And
        [PRED\_1]
        [PRED\_2]
    ]
]
\end{forest}
\end{center}
The corresponding preorder index is [Exist:0, And:1, PRED\_1:2, PRED\_2:3], which means Persephone can use a length-four vector to completely record what happened during a game played on this tree. All it needs is storing the action taken on each node to the corresponding position in the vector. For the given instance above, suppose we choose $m=5$ and then take the left branch (marked as 0 in action space) of the ``And" formula, we will end up with a vector [5,0,-1,-1] at PRED\_1, where -1 is the default value for null actions. After combining the input parameters and the identification of the predicate, a complete state representation has the form:
$$[PRED\_ID,PARAM\_LIST,TREE\_PREORDER]$$
Backing to the previous example, suppose $k=q=4$ and $n=16$, we have:
$$[PRED\_ID,4,4,16,5,0,-1,-1]$$
when the game stops at $PRED\_1:HSR(4-1,4-1,5)$. 
\subsubsection{Warm-start MCTS}
It is known that MCTS cannot handle large action space because of the sample efficiency problem. To be specific, think of a game tree with a significant branching factor while the winning leaf nodes are incredibly sparse. In this case, MCTS can find the winning strategy if and only if it can traverse as many paths as possible, which requires a large number of MCTS simulations. The number of simulations soon becomes intractable when the action space increases. To mitigate this issue,  We applied the warmstart trick.

The idea is quite simple: in the original design of neural MCTS, the search tree is always re-initialized before each MCTS simulation session. That means MCTS has to recount the visiting count for each node from the beginning. When the action space increases, such recounting makes it impossible to locate the optimal path with a relatively small number of simulations. Therefore, we propose to use a warm-start MCTS to accelerate the simulation process. The warm-start mainly contains the following two components: 
\begin{enumerate}
    \item Keeping counting info: after each MCTS simulation session, we will keep the node's counting $N(s,a)$ in the search tree and reuse it in the next iteration of the simulation session. We will see in the experiment that a substantial learning speed improvement by merely applying this trick.
    \item Q-value injection: inspired by the idea from SAVE \cite{Hamrick2020CombiningQA}, where the author proposes a variant of neural MCTS which only learns the Q-values and uses a softmax operator to recover the policy from the learned Q-values. We inject the predicted Q-values into the search tree nodes as priors, and our experiment shows that it will make the learning process more stable. 
\end{enumerate}
\subsubsection{Policy Learning}
The neural MCTS can be viewed as an Actor-Critic (AC) algorithm \cite{ACSurvey}. The self-play phase works like an actor which optimized empirical policy $\hat{\pi}$ in the direction suggested by the value network, the Critic, then samples game trajectories from the optimized empirical policy $\hat{\pi}$. In the training phase, the Actor will update the policy network towards the empirical policy $\hat{\pi}$; meanwhile, the Critic will update the value network using the reward signal from the sampled trajectories.

An AC interpretation of neural MCTS makes alternative policy learning approaches becomes feasible. Specifically, in the original AlphaZero implementation, policy network is updated through a cross entropy method (CEM,\cite{Mannor2003TheCE}), where the policy loss is defined as:
$$\mathcal{L}(\theta)=-\sum_t\hat{\pi}(s_t,a_t)log \pi_{\theta}(s_t,a_t)$$
However, further research in policy gradient suggests that using more sophisticated methods like TRPO and PPO can improve the stability of the learning process \cite{ppo,trpo}. Therefore, we also implemented the PPO policy update as alternatives. In our experiment, we mainly focus on the following two PPO policy loss variants:
\begin{equation}
\begin{split}
    \mathcal{L}^{CLIP}(\theta)&=\mathbb{E}_t\left[\min(r(\theta)A(s_t,a_t), \text{clip}(r(\theta)A(s_t,a_t),1-\epsilon,1+\epsilon))\right]\\
    \mathcal{L}^{KL}(\theta)&=\mathbb{E}_t\left[r(\theta)A(s_t,a_t)-\beta KL[\hat{\pi},\pi_\theta]\right]
\end{split}
    \label{eq:ppo-loss}
\end{equation}
where the ratio $r(\theta)=\frac{\pi_{\theta}(s_t,a_t)}{\hat{\pi}(s_t,a_t)}$ is the importance weights, and $A(s,a)$ is the advantage term, which is defined as:
\begin{equation}
\begin{split}
    A(s,a)=\begin{cases} 
    -V_\theta(s)-r,\ \mbox{if } \mathcal{T}(s,a) \mbox{ ends the game with reward }r \\ 
    -V_\theta(\mathcal{T}(s,a))-V_\theta(s),\ \mbox{otherwise}
\end{cases}
\end{split}
    \label{eq:ppo-adv1}
\end{equation}
when the next player is not the current player. And
\begin{equation}
\begin{split}
    A(s,a)=\begin{cases} 
    -V_\theta(s)+r,\ \mbox{if } \mathcal{T}(s,a) \mbox{ ends the game with reward }r \\ 
    V_\theta(\mathcal{T}(s,a))-V_\theta(s),\ \mbox{otherwise}
\end{cases}
\end{split}
    \label{eq:ppo-adv2}
\end{equation}
when the next player is the current player.
\subsubsection{Multiple Neural Networks}
There are mainly two reasons why we consider multiple neural networks in our implementation:
\begin{enumerate}
    \item As we have mentioned earlier that a semantic game can be an asymmetric extensive form game, which means the two players learns different value and policy networks. Therefore, it is useful to know whether to use separate neural networks for each player would be helpful on their learning efficiency. For instance, in the $HSR(k,q,n)$ semantic game, the Proponent player has a policy of size $n$ while the Opponent only needs to choose from two actions. In \cite{Xu2019LearningSA}, the author shows a behavior asymmetry on playing those games, which might indicate a performance improving if the two players learn on different neural networks.
    \item As suggested in \cite{Andrychowicz2020WhatMI}, the author shows experimentally that separate value and policy networks can generally increase the learning performance. Even though AlphaZero is prone to integrate the two into one neural network, it might be a bad idea because the training signal can interfere with each other and cause the learning process to become unstable. In our experiment, we also test this idea and justify the observation in \cite{Andrychowicz2020WhatMI}. 
\end{enumerate}

\section{Experimental Evaluation}
\subsection{Experiment Setup}
\subsubsection{Target Problem}
We test our idea on the HSR problem, which has been introduced and also used as experimental subjects in \cite{Xu2019LearningSA}. $HSR(k,q,n)$ basically defines a stress testing problem, where one, given $k$ jars and $q$ test chances, throwing jars from a specific rung of a given ladder with height $n$ to locate the highest safe rung. If $n$ is appropriately large, then one can locate the highest safe rung with at most $k$ jars and $q$ test times; otherwise, if $n$ is too big, then there is no way to locate the highest safe rung. This problem can be described with the following FOL:
$$HSR(k,q,n)=\begin{cases} \mbox{True},\ \mbox{if } n=1 \\ \mbox{False},\ \mbox{if } n>1\wedge (k=0\vee q=0) 
\\\exists m\in [1,n):
HSR(k-1,q-1,m)\\\wedge HSR(k,q-1,n-m),\ \mbox{otherwise} 
\end{cases}$$
Moreover, we perform our experiment mainly on $HSR(7,7,128)$ in the following context, if we do not specify other instances.
\subsubsection{Neural Network Setup}
\begin{itemize}
    \item For the single neural network, we use a two-head MLP for both policy and value network. The shared layers has shape $[1024,1024,1024,512]$, with ReLu activation. Them, for the policy head, it has shape $[|A|_max]$ (where $|A|_max$ is the maximum size of action space), with Softmax activation. For the value head, it has shape $[1]$, with Tanh activation.
    \item For the separated policy and value neural networks, we just split the shared layers into a different neural network, so that they are independent with each other.
    \item For the separated player neural network, to respect the game's asymmetry, we use $[1024,1024,1024,512]$ for the Proponent network, while $[256,256,256,128]$ for the Opponent network. It is to be noted that separate player networks is independent of separate policy/value networks, which means one can have separated player networks while each player still uses a single neural network or uses a separated policy/value networks.
\end{itemize}
\subsubsection{Experimental Configurations}
Since there are multiple implementations to compare, we define each implementation setup here for later usage.
\begin{itemize}
    \item AZ: the original implementation of AlphaZero.
    \item CE: the keep-counting-info warm-start variant of AZ.
    \item CE\_Sep: the separate policy/value network variant of CE.
    \item CE\_Q\_Sep: the full warm-start (i.e., keeping-counting-info and Q-value injection) variant of CE\_Sep.
    \item PPO\_CLIP\_Sep: the separate policy/value network with PPO policy learning, which uses the $\mathcal{L}^{CLIP}$ loss.
    \item PPO\_KL\_Sep: the separate policy/value network with PPO policy learning, which uses the $\mathcal{L}^{KL}$ loss.
    \item PPO\_KL\_Sep\_2NN: the separate player network variant of PPO\_KL\_Sep.
\end{itemize}
\subsubsection{Hyperparameters}
\begin{itemize}
    \item Learning rate to 0.001 with Adam optimizer.
    \item The mini-batch size is set to 64.
    \item Training epochs is set to 10.
    \item Number of MCTS simulation is set to 25.
    \item number of self-play is set to 100.
    \item Replay buffer limit is set to 20 iterations.
    \item $\beta$ for $\mathcal{L}^{KL}$ is 1.
    \item $\epsilon$ for $\mathcal{L}^{CLIP}$ is 0.2.
\end{itemize} 
\subsection{Evaluation Methodology}
The evaluation phase is like the self-play phase, where the players optimize their moves through an MCTS simulation to play against each other. However, instead of using the same neural networks for both players, we use the newly trained policy/value networks to play against the networks from the last iteration. Specifically, we first run a given number (in our experiment, it is 20) of games between the Proponent, who uses the newly trained networks, and the Opponent, who uses the previously trained networks; then we run the same number of games between the Proponent, who uses the previously trained networks, and the Opponent, who uses the newly trained networks.

We then use fault counting to measure the performance of each game. The concept of fault counting is based on the correctness measurement in \cite{Xu2019LearningSA}, where the action is correct if it preserves a winning position:
\begin{itemize}
    \item Proponent's correctness: Given $(k,q,n)$, correct actions exist only if $n\le N(k,q)$. In this case, all testing points in the range $[n-N(k,q-1),N(k-1,q-1)]$ are acceptable. Otherwise, there is no corrective action.
    \item Opponent's correctness: Given $(q,k,n,m)$, When $n>N(k,q)$, any action is regarded as correct if $N(k-1,q-1)\le m\le n-N(k,q-1)$, otherwise, the OP should take ``not break'' if $m>n-N(k,q-1)$ and ``break' if $m<N(k-1,q-1)$; when $n\le N(k,q)$, the OP should take the action ``not break'' if $m<n-N(k,q-1)$ and take action ``break'' if $m>N(k-1,q-1)$. Otherwise, there is no corrective action.
\end{itemize}
Fault counting then counts when a player makes a mistake that there is a move to keep the winning position, but the player chooses an incorrect one. If the player makes a mistake, while the opponent player catches that mistake by moving to a winning position, we increase the fault counting by one for the player who makes that mistake. It is to be noted that a player can lose the game without making any fault. That is because the player is in a losing position, and there is no corrective action to take. In other words, if a player is forced to lose, then he should not be blamed for losing that game.

The learning is considered to be converged if both newly trained and previously trained networks show zero faults for a certain number (in our experiment, 5) of consecutive learning iterations. In our experiment, we measure the fault counting and the number of iterations needed before convergence for each configuration, along with the loss curve, we can evaluate the performance of different configuration.

\subsection{Results and Discussion}
\subsubsection{Does keep-counting-info help?} 
\begin{figure}[h]
  \centering
  \includegraphics[width=\linewidth]{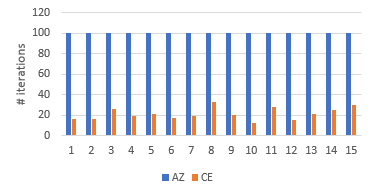}
  \caption{Number of iterations needs before the first time both of the players got 0 faults. It can be seen that without keep-counting-info, AZ even cannot find the optimal policy by any chance. Actually, AZ can find the optimal policy only when we increase MCTS simulation times to 100, which is four times larger than the current value.}
  \label{fig:AZ-CE}
  \Description{Number of iterations needs before the first time both of the players got 0 faults. It can be seen that without keep-counting-info, AZ even cannot find the optimal policy by any chance. Actually, AZ can find the optimal policy only when we increase MCTS simulation times to 100, which is four times larger than the current value.}
\end{figure}

In this experiment, we measure the performance of AZ and CE. We set the maximum number of iteration to 100 and then measure the number of iterations needed before fault counting first time arrives 0 for both of the players. We run the experiment for 15 times, and it has been shown in Fig.\ref{fig:AZ-CE} that CE converges in 35 iterations, while AZ never converges, given the hyperparameters we have mentioned in the previous section. The experimental result justified that merely adding keep-counting-info warm-start can already improve the efficiency significantly.
\subsubsection{Does separate policy/value networks help?} 
\begin{figure}[h]
  \centering
  \includegraphics[width=\linewidth]{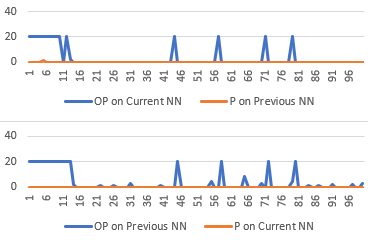}
  \caption{The fault-counting measurement from one of the 15 experiments on CE. We only show one experiment because of page limits, but all 15 experiments have a similar curve where spikes appear after convergence, which signifies unstable learning.}
  \label{fig:CE_1}
  \Description{The fault-counting measurement from one of the 15 experiments on CE. We only show one experiment because of page limits, but all 15 experiments have a similar curve where spikes appear after convergence, which signifies unstable learning.}
\end{figure}

\begin{figure}[h]
  \centering
  \includegraphics[width=\linewidth]{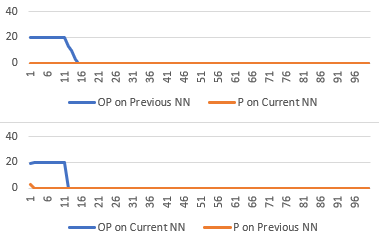}
  \caption{The fault-counting measurement from one of the 15 experiments on CE\_Sep. Because of page limits, we only show one experiment, but all 15 experiments have a similar curve where no spike appears after convergence, which is a sign of stable learning.}
  \label{fig:CE_2}
  \Description{The fault-counting measurement from one of the 15 experiments on CE\_Sep. Because of page limits, we only show one experiment, but all 15 experiments have a similar curve where no spike appears after convergence, which is a sign of stable learning.}
\end{figure}

In this experiment, we measure the performance of CE and CE\_Sep. We set the maximum number of iteration to 100 and then measure the fault-counting after each iteration. We run the experiment 15 times, and found that CE can be unstable compared to CE\_Sep (as shown in Fig.\ref{fig:CE_1} and Fig.\ref{fig:CE_2}). This verified the observation from \cite{Andrychowicz2020WhatMI}, where the author also recommended to use a separated policy/value neural network. We think those spikes are due to interfered target signals in a layer-shared neural network, where value signal and policy signal interfere with each other inevitably but unnecessarily.
\subsubsection{Does Q-value injection/PPO/separated player networks help?}
\begin{figure}[h]
  \centering
  \includegraphics[width=\linewidth]{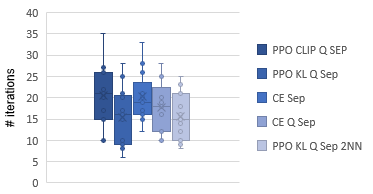}
  \caption{The Box and Whisker graph for the number of iterations needed before convergence for different configurations.}
  \label{fig:REST_BOX}
  \Description{The Box and Whisker graph for the number of iterations needed before convergence for different configurations.}
\end{figure}

Since keep-counting-info and separate policy/value networks are essential for efficiency and stability, we always apply these two configurations in the rest of our experiment. In this experiment, we measure the average number of iterations need before convergence for CE\_Sep, CE\_Q\_Sep, PPO\_CLIP\_Sep, PPO\_KL\_Sep, and PPO\_KL\_Sep\_2NN. We run 20 experiments on each configuration and plot the Box and Whisker graph for each of them (see Fig.\ref{fig:REST_BOX}).

Our experiment provides us several meaningful results:
\begin{enumerate}
\item Q-value injection helps increase the efficiency of the algorithm. It can be seen from the graph that CE\_Q\_Sep averagely locates the optimal policy faster than CE\_Sep. That is because Q-values from the value network accelerates the search process of MCTS, which helps it evaluate the UCB formula unbiased. As a result, we assume Q-value injection in the rest of our experiments.
\item Using $\mathcal{L}^{KL}$ is much better than using $\mathcal{L}^{CLIP}$. As shown in the graph, PPO\_CLIP\_Sep has a higher average value and a larger variance than PPO\_KL\_Sep. We think this might be due to the fact of reward sparsity in semantic games, which will cause the advantage to become very small hence provide less information to update the policy network for $\mathcal{L}^{CLIP}$. On the other hand, $\mathcal{L}^{KL}$ has a KL regularization term, which works like cross-entropy and provides more information for learning the policy network.
\item It seems that separate player neural networks perform slightly worse than using the same policy/value network for both of the players. It is to be noted that PPO\_KL\_Sep\_2NN runs much faster than PPO\_KL\_Sep because of the smaller network size for the Opponent player. It takes averagely 10s to finish one training epoch for PPO\_KL\_Sep, while only 2s for PPO\_KL\_Sep\_2NN. Therefore, even though\\ PPO\_KL\_Sep\_2NN needs averagely 2 to 3 more iterations to find the optimal strategy, it still takes less time than PPO\_KL\_Sep, which means separate player neural networks do increase efficiency.
\end{enumerate}

\begin{figure}[h]
  \centering
  \includegraphics[width=\linewidth]{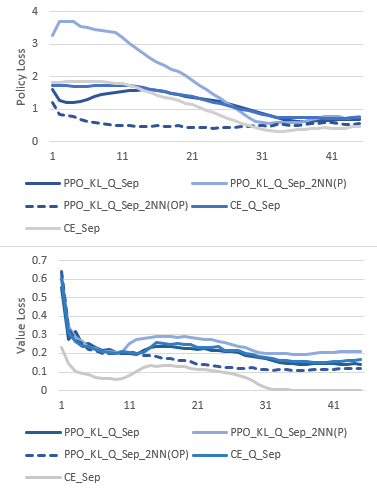}
  \caption{The loss curve for different configurations. Notice that we did not plot PPO\_CLIP\_Q\_Sep, because the loss curve of $\mathcal{L}^{CLIP}$ behaves differently, which makes it incomparable to the rest of the loss curves.}
  \label{fig:Loss}
  \Description{The loss curve for different configurations.}
\end{figure}

To further investigate the behavior of different configurations, we also plot the loss curve for both policy and value network for each configuration (Fig. \ref{fig:Loss}). CE's value loss curve shows that it drops much faster than other configurations, which means that the value network might quickly converge to local optimal at the beginning of learning. Hence it takes a longer time to jump out of the local optimal and lower the efficiency. Another fact to notice is that the policy loss of Opponent's networks in the PPO\_KL\_Q\_Sep\_2NN configuration converges earlier than Proponent's networks, which means the separated player networks captures the asymmetry properly in this semantic game.     

\subsection{Performance Scoring}
\begin{figure}[h]
  \centering
  \includegraphics[width=\linewidth]{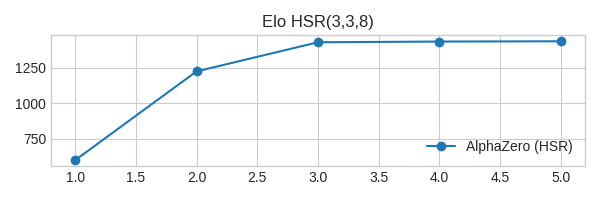}
  \includegraphics[width=\linewidth]{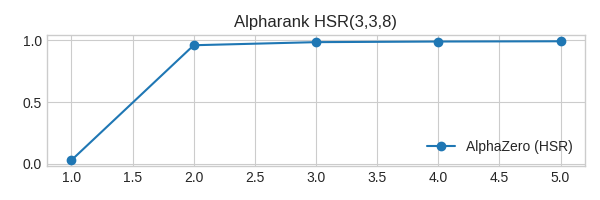}
  \caption{For $HSR(3,3,8)$, the convergence happened very quickly when the instance is small. The $\alpha$-Rank score also shows a very similar pattern. A further checking on the fault-counting metric confirmed that both players have converged to 0 faults.}
  \label{fig:elo_1}
  \Description{This figure shows the Elo-rating and the $\alpha$-Rank score for HSR(3,3,8) configuration as the training progresses}
\end{figure}

\begin{figure}[h]
  \centering
  \includegraphics[width=\linewidth]{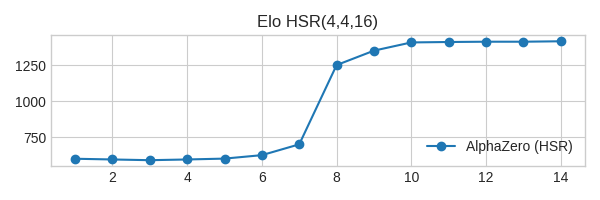}
  \includegraphics[width=\linewidth]{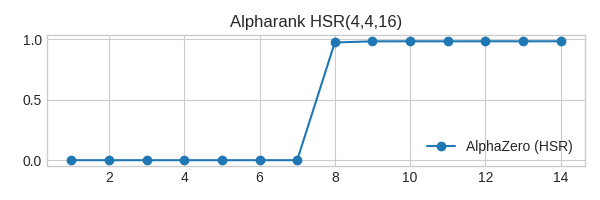}
  \caption{For $HSR(4,4,16)$, the Elo-rating starts at 600 and drops until iteration 4, after the fourth iteration the increase in the Elo-rating is almost linear up to iteration 7 after which there is a huge jump of around 550 points to iteration 8 and a 220 point jump to iteration 9  stays constant at around 1420 thereafter. The $\alpha$-Rank score for this experiment stays constant at 0.02 for the first 7 steps and again shoots to 0.97 at iteration 8 and remains constant at 0.99 thereafter.}
  \label{fig:elo_2}
  \Description{This figure shows the Elo-rating and the $\alpha$-Rank score for HSR(4,4,16) configuration as the training progresses}
\end{figure}

It is to be noted that HSR is a special problem for which the ground-truth is already known. However, in general, for most combinatorial problems, the ground-truth is unknown, requiring us to figure out an alternative to measure the performance. In this additional experiment, we use two performance scoring techniques, Elo-rating \cite{elorating} and $\alpha$-Rank \cite{alpharank}, to score the Proponent player per each learning iteration. Performance scoring allows us to measure a player's performance simply through the game results or the pay-off table from the competition with other player instances. For Elo-rating, we run a competition between the two players after each training iteration and compute their score; For $\alpha$-Rank, we have to store all player instances after each iteration and run a competition among different players from different training iterations to generate a pay-off table, then we run the $\alpha$-Rank algorithm to get the score. In order to generate the pay-off table more efficiently, we run the scoring process on two relatively smaller instances: $HSR(3,3,8)$ (Fig.\ref{fig:elo_1}) and $HSR(4,4,16)$ (Fig.\ref{fig:elo_2}). We can see a clear phase transition where the algorithm jumps from a low score to a high score, which indicates an optimal strategy has been found.

\section{Conclusion}
This paper proposed a framework, Persephone, to efficiently map a first-order problem to a two-player semantic game and then play and learn an optimal game strategy through a neural MCTS based RL algorithm. The optimal learned strategy can then be mapped back to an optimal solution for the original problem. We also introduced a formal definition for symmetric/asymmetric extended form games, which motivates us to investigate asymmetric neural network designs. We proposed several variants to the vanilla AlphaZero algorithm, such as using different policy learning strategies, warm-start, separate policy/value networks, separate player networks, and carried out experiments on different configurations. The experimental results can be measured either through ground-truth based metrics, like fault-counting in our case or through performance scoring techniques like Elo-rating and $\alpha$-Rank. Our experimental results show that a KL-divergence regularized PPO policy learning with warm-start MCTS and separated neural networks perform the best, which justified our improvements to the original AlphaZero algorithm.






\bibliographystyle{ACM-Reference-Format} 
\bibliography{sample}


\end{document}